%% file: main.tex
\documentclass[10pt,twocolumn,letterpaper]{article}

\usepackage{cvpr}              % To produce the CAMERA-READY version
\usepackage[table]{xcolor}
\usepackage{multirow}
\usepackage{pifont}
\usepackage{tcolorbox}
\usepackage{bm}
\usepackage{float}
\usepackage{algorithm}
\usepackage{algorithmic}
\usepackage{rotating}
\usepackage{bbm}
\definecolor{cvprblue}{rgb}{0.21,0.49,0.74}
\usepackage[pagebackref,breaklinks,colorlinks,allcolors=cvprblue]{hyperref}

\def\paperID{18377} % *** Enter the Paper ID here
\def\confName{CVPR}
\def\confYear{2025}

\title{Evolution of Thought: Diverse and High-Quality Reasoning via Multi-Objective Optimization}

\author{
  Biqing Qi\textsuperscript{1}, 
  Zhouyi Qian\textsuperscript{2},
  Yiang Luo\textsuperscript{3},
  Junqi Gao\textsuperscript{1,2,}\thanks{Corresponding authors.},
  Dong Li\textsuperscript{1,2},
  Kaiyan Zhang\textsuperscript{4}, 
  Bowen Zhou\textsuperscript{1,4,}\footnotemark[1] \\
  $^1$ Shanghai Artificial Intelligence Laboratory, \\
  $^2$ School of Mathematics, Harbin Institute of Technology, \\
  $^3$ Department of Control Science and Engineering, Harbin Institute of Technology, \\
  $^4$ Department of Electronic Engineering, Tsinghua University \\
  {\tt\small \{qibiqing7,qianzhouyi0,normanluo668,gjunqi97,arvinlee826\}@gmail.com,} \\ 
  {\tt\small zhang-ky22@mails.tsinghua.edu.cn,zhoubowen@tsinghua.edu.cn}
  }

\begin{document}
\newtcolorbox{mycolortbox}{
    colback=blue!5, 
    colframe=black, 
    arc=4pt, 
    boxrule=1pt, 
    fonttitle=\bfseries
}
\maketitle
\input{sec/0_abstract}
\input{sec/1_intro}

\input{sec/2_Related_works}
\input{sec/3_Methodology}
\input{sec/4_Experiments}

\input{sec/5_Conclusion}
\clearpage
{
    \small
    \bibliographystyle{ieeenat_fullname}
    \bibliography{main}
}
\input{sec/X_suppl}
% WARNING: do not forget to delete the supplementary pages from your submission 
% \input{sec/X_suppl}

\end{document}

%% file: sec/0_abstract.tex
\begin{abstract}

As multi-modal large language models (MLLMs) are increasingly applied to complex reasoning tasks, the diversity and quality of reasoning paths become crucial factors affecting their performance. Although current methods aim to enhance reasoning quality through path expansion, they often neglect the diversity of reasoning paths and effective information sharing, leading to local optima and inefficiency.
To address these challenges, we propose Evolution of Thought (EoT), a multi-objective framework designed to improve reasoning by fostering both high-quality and diverse reasoning paths.
Specifically, we introduce the Non-dominated Sorting Genetic Algorithm II for multi-objective optimization, utilizing crossover and mutation operators to promote greater diversity in reasoning solutions. Additionally, we propose a Condensation-Aggregation mechanism to cluster and eliminate redundant paths, facilitate improved information sharing among parent nodes, and ultimately enhance both the efficiency and quality of the reasoning process.
Validation experiments on various vision-language and language reasoning tasks demonstrate that EoT achieves superior reasoning performance and efficiency compared to other competitive baselines. Our study provides a novel perspective on the design of heuristic reasoning frameworks for MLLMs.

\vspace{-5pt}

\end{abstract}

%% file: sec/1_intro.tex
\section{Introduction}
\label{sec:intro}

\begin{figure}[t]
    \centering
    \includegraphics[width=0.43\textwidth]{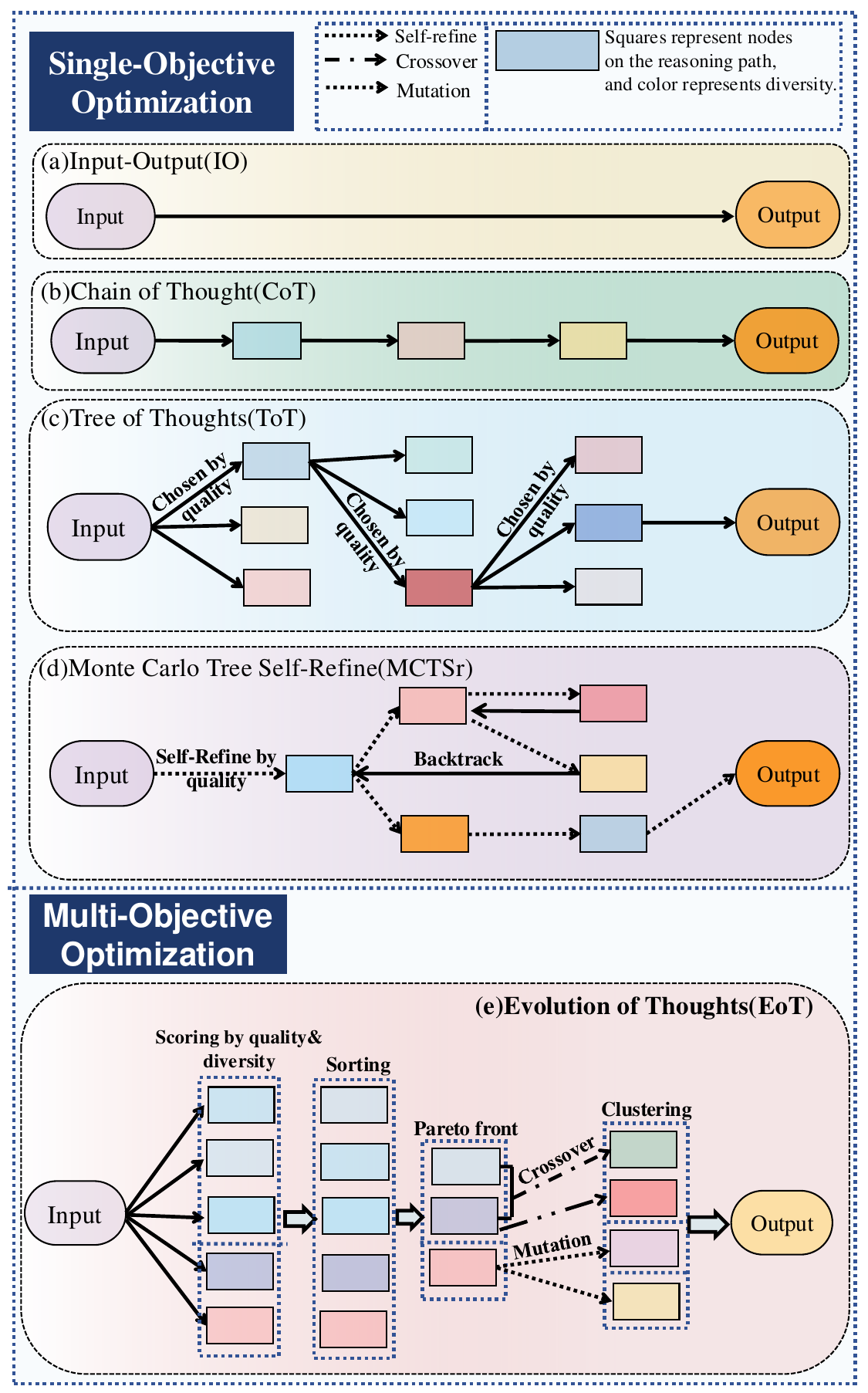}
    \caption{Comparison of Evolution of Thoughts (EoT) with other approaches. Color depth indicates answer quality: blue for high quality, red for low quality.}
    \label{fig:Comparison}
    \vspace{-15pt}
\end{figure}

Multi-modal large language models (MLLMs) are widely adopted due to their generalization capabilities and exceptional performance across various tasks \cite{yang2024unveiling}. However, MLLMs often face challenges in achieving accuracy and consistency in open-domain question answering, complex mathematical reasoning, and tasks requiring strict logical reasoning \cite{liang2024internal, ahn2024large}. This is primarily due to their reliance on a 'black-box reasoning approach' \cite{kim2024learning}. Typically, these models generate responses in a single step, directly based on the input, without sufficient control over the reasoning process needed to maintain logical coherence in complex tasks \cite{Wei0SBIXCLZ22}. As a result, the generated answers often suffer from causal ambiguity or internal inconsistencies. In tasks requiring multiple sequential reasoning steps, errors can accumulate, leading to a significant degradation in the quality of the final output. To address this limitation, it is crucial to introduce an explicit reasoning mechanism that enables the model to progress through the reasoning process step by step, thereby improving both accuracy and logical consistency.

To address these challenges, \textbf{C}hain \textbf{o}f \textbf{T}houghts (CoT) \cite{Wei0SBIXCLZ22} emerges as a promising solution. By explicitly breaking down the reasoning process into multiple logical steps, CoT not only significantly enhances the model's reasoning capabilities in highly uncertain contexts but also improves its adaptability and generalization to complex tasks through a multi-step reasoning strategy. However, CoT relies on a greedy decoding (top-1) strategy along a single path, selecting only the optimal path at each step, which restricts the diversity of reasoning paths. This design limits the model's ability to explore alternative answer paths in complex tasks, thereby affecting both the quality of solutions and completion reliability.

To overcome the single-path limitation of CoT, \textbf{T}ree \textbf{o}f \textbf{T}houghts (ToT) \cite{YaoYZS00N23} has been proposed. ToT enables the model to explore multiple reasoning paths concurrently using depth- and breadth-first search within a tree structure, increasing the probability of finding the optimal solution. Expanding upon ToT, the \textbf{G}raph \textbf{o}f \textbf{T}houghts (GoT) \cite{BestaBKGPGGLNNH24} introduces node aggregation, integrating a graph structure information into the search process to allow the model to leverage prior reasoning steps, thereby enhancing generation quality. However, GoT relies on predefined aggregation operations, making it better suited to well-structured, fixed-state problems (e.g., numerical sorting) and less flexible for complex, dynamic reasoning tasks. Moreover, both ToT and GoT experience rapid search path expansion with increasing task complexity, leading to lower reasoning efficiency. To address the challenge of search efficiency, the Monte Carlo Tree Self-refine (MCTSr) method \cite{abs-2406-07394} leverages dynamic pruning strategies to improve reasoning efficiency. Although MCTSr increases both path diversity and search efficiency to some extent, it tends to neglect diversity when optimizing reasoning paths makes it prone to converging to local optima. As paths are expanded, the model’s generated paths tend to converge on a few high-scoring paths, restricting the method’s overall global exploration.

In summary, two major challenges remain in current reasoning path optimization: \noindent\textbf{1) Convergence to Local Optima.} Although ToT and MCTSr improve path diversity, their optimization primarily focuses on generation quality, which often causes paths to converge on high-scoring branches, thereby limiting diversity and the achievement of global optima \cite{zhao2024many}; \noindent\textbf{2) Information Sharing and Search Efficiency Limitations.} Current methods often lack mechanisms for effectively sharing information from parent nodes, reducing search efficiency. This lack of information reuse from parent nodes hinders search efficiency during tree expansion. Although GoT introduces an aggregation mechanism, it relies on predefined operations, limiting its flexibility and broader applicability.

To address these limitations, we introduce the \textbf{E}volution \textbf{o}f \textbf{T}houghts (EoT) framework for enhancing reasoning. EoT formulates reasoning as a multi-objective optimization (MOO) problem, balancing both answer quality and diversity. To solve this MOO effectively, we introduce the Non-dominated Sorting Genetic Algorithm II (NSGA-II) \cite{996017}, which utilizes non-dominated sorting to optimize both diversity and quality. By maintaining a diverse set of solutions and preventing convergence to narrow thought patterns, NSGA-II mitigates the tendency of traditional genetic algorithms to become trapped in local optima, thus enhancing the overall optimization process. To prevent redundant paths in NSGA-II’s search process, we propose a Condensation-Aggregation (CA) mechanism that clusters paths to reduce redundancy and enhances information sharing for improved reasoning efficiency. The framework operates in three main steps: \textbf{1) Quality and diversity assessment. }Using MLLMs to score generated answers and construct a quality metric, while defining a novelty metric based on semantic similarity, forming a MOO problem that balances quality and diversity; \textbf{2) Multi-objective sorting. }Utilizing non-dominated sorting to rank solutions by quality and diversity across different dominance layers, selecting high-quality solutions on the Pareto frontier as the new parent generation; \textbf{3) Crossover and mutation. }Designing crossover and mutation strategies that aggregate information from parent generations and generate diverse new candidate solutions, enhancing answer diversity.

Finally, through the CA mechanism, EoT removes redundant information while ensuring answer quality and diversity, thereby enhancing the summarization efficiency and stability of MLLMs. Experimental results show that EoT outperforms existing baselines on various vision-language and language-only reasoning tasks.

Therefore, our contributions can be outlined as follows:

\begin{itemize}
\item We extend reasoning path search from a single-objective focus on answer quality to MOO, improving generalization and adaptability by considering both quality and diversity.
\item Building on MOO, we propose EoT, which uses NSGA-II to optimize both answer quality and diversity, generating diverse candidate solutions while preserving quality.
\item To reduce redundancy in NSGA-II's optimization, we propose the CA mechanism, which clusters paths to eliminate redundancy, enhancing reasoning efficiency and information flow, and offering more flexibility than GoT.

\end{itemize}

%% file: sec/2_Related_works.tex
\vspace{-5pt}
\section{Related Works}
\label{sec:Related Works}
\begin{figure*} [t]
	\centering
\includegraphics[width=0.9\textwidth]{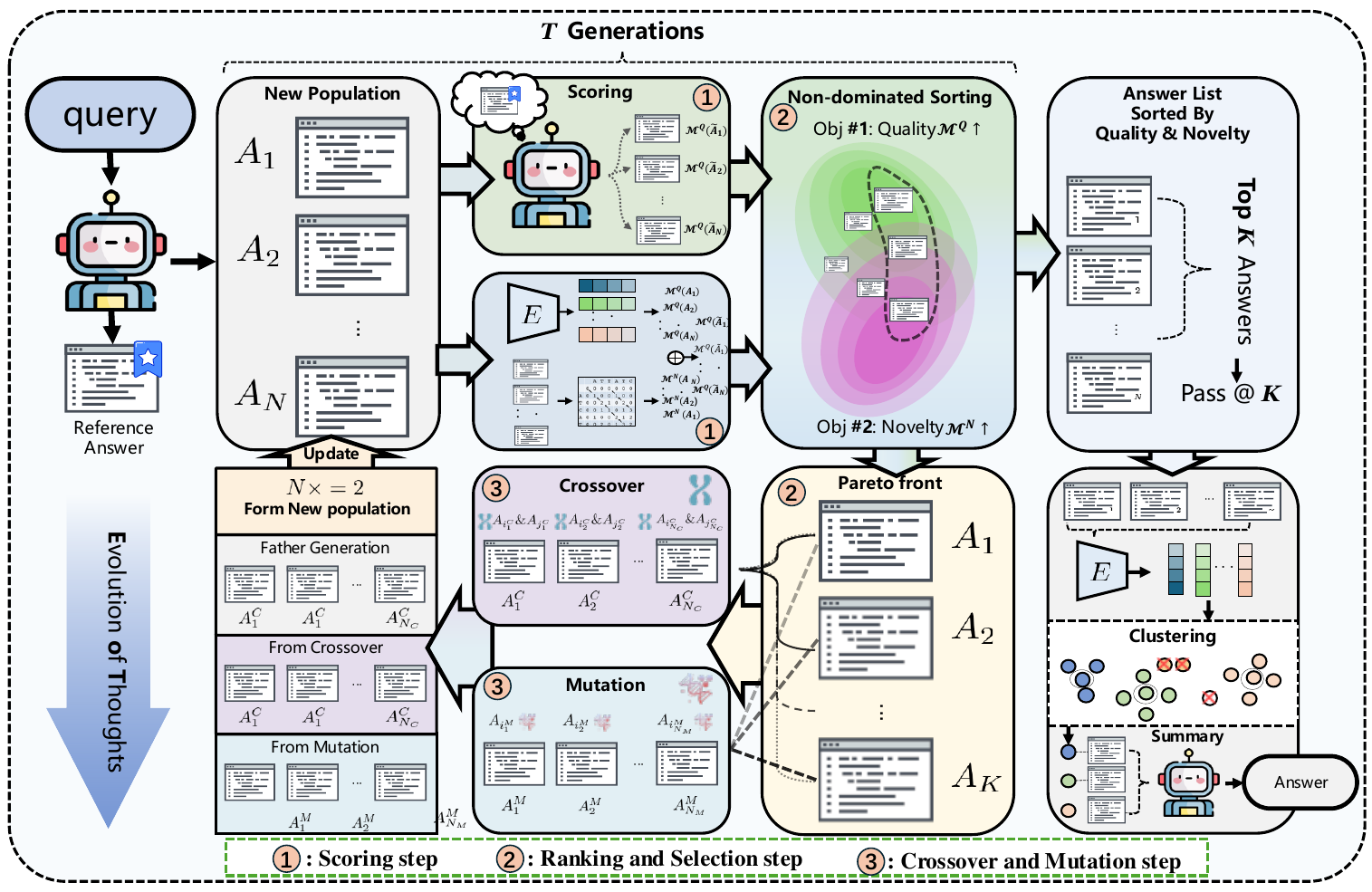}
	\caption{EoT Process Flow Diagram.Step 1 involves scoring using $\mathcal{M}^{Q}(A)$ and $\mathcal{M}^{N}(A)$; Step 2 involves non-dominated sorting; Step 3 includes Crossover and Mutation operations to generate diverse offspring.}
	\label{EoT_Diagram} 
\vspace{-15pt}
\end{figure*}

\noindent\textbf{LLMs Reasoning.} Recent research has made significant progress in enhancing LLM and MLLM generation quality through prompt engineering, optimizing these models for diverse downstream tasks. CoT \cite{Wei0SBIXCLZ22} guides LLMs and MLLMs to generate a series of intermediate steps leading to the final answer. By breaking down the problem-solving process, CoT simplifies each individual step and provides a new approach for tackling complex reasoning tasks. To overcome the single-path limitation of CoT, ToT \cite{YaoYZS00N23} is proposed. ToT allows the model to explore multiple reasoning paths concurrently using depth- and breadth-first search within a tree structure, increasing the likelihood of finding the optimal solution. Building on ToT, GoT \cite{BestaBKGPGGLNNH24} introduces node aggregation, integrating a graph structure into the search process to enable the model to leverage prior reasoning steps, thereby enhancing generation quality. However, GoT relies on predefined aggregation operations, making it better suited for well-structured, fixed-state problems (e.g., numerical sorting) and less flexible for dynamic, complex reasoning tasks. Moreover, both ToT and GoT suffer from rapid search path expansion as task complexity increases, leading to reduced reasoning efficiency. To address the challenge of search efficiency, MCTSr \cite{abs-2406-07394} employs dynamic pruning strategies to improve reasoning efficiency. Although MCTSr increases both path diversity and search efficiency to some extent, it still tends to converge toward local optima. As paths expand, the model’s generated paths tend to converge on a few high-scoring paths, limiting the method’s global exploration. Subsequent works have primarily focused on improving answer refinement \cite{Self-refine, shinn2024reflexion, yao2022react, paul2023refiner}, multi-perspective interactive reasoning \cite{liang2023encouraging, wang2023unleashing, wang2024mixture, du2023improving}, and reasoning structures \cite{abs-2406-07394, YaoYZS00N23, BestaBKGPGGLNNH24}. Methods such as Self-Refine \cite{Self-refine}, Reflexion \cite{shinn2024reflexion}, and Refiner \cite{paul2023refiner} leverage model feedback by evaluating the generated responses and selecting appropriate optimization directions based on this feedback to improve reasoning quality. However, due to an excessive emphasis on thorough deliberation, these methods suffer from low computational efficiency \cite{swiechowski2023monte}.

\noindent\textbf{Heuristic Algorithms. }Recently, tree search \cite{abs-2406-07394,YaoYZS00N23,chen2024alphamath,xu2023no} and graph search \cite{BestaBKGPGGLNNH24}, have been widely applied in LLMs and MLLMs reasoning and have achieved significant success. These methods significantly enhance the reasoning capabilities of LLMs and MLLMs, particularly in complex tasks like mathematical problem-solving, by integrating Monte Carlo Tree Search with fine-tuned LLMs \cite{chen2024alphamath}, ranking decision steps for immediate responses and precise reasoning \cite{abs-2406-07394,xu2023no}, and incorporating aggregation operations with graph structures to allow new nodes to accumulate prior information \cite{BestaBKGPGGLNNH24}. However, these approaches face limitations: 1) relying solely on answer quality for search and expansion can result in rigid reasoning, and 2) inefficient searches due to the inability to share information across parent nodes. Genetic algorithms, with their population-based selection, crossover, and mutation operations, offer a promising solution to these issues. Concepts such as non-dominated sorting \cite{schaffer1985multiobjective}, crowding distance, and elitism strategies \cite{deb2002fast} have further refined genetic algorithms, which are already widely recognized in the neural network domain \cite{elsken2019neural}. However, these methods remain underexplored in the context of LLM reasoning, specially MLLM. Optimizing the search direction while maintaining diversity in the search space to improve both reasoning quality and solution diversity remains an open question.

%% file: sec/3_Methodology.tex
\section{The Framework of EoT}

In this section, we introduce the EoT framework, depicted in Figure \ref{EoT_Diagram}, and describe its core components and workflow in detail. We begin by defining the reasoning enhancement process as a MOO problem, with two primary objectives: answer quality and diversity. Next, we design the evolutionary search process of EoT and define each operation to optimize the search for high-quality and diverse answers. Finally, we introduce the CA mechanism to help the model refine, summarize, and provide reliable responses.
\subsection{Problem Formulation}

Let the user query prompt be denoted as $\bm{p}_{q}$, and let the MLLM (or LLM) be represented by $f$. Given an evaluation criterion $\mathcal{M}$, where higher values of $\mathcal{M}$ correspond to better performance, the objective is to design a search strategy $\mathcal{S}$ that optimizes the model's output based on $\mathcal{M}$. Specifically, the goal is to maximize $\mathcal{M}(\mathcal{S}(f, \bm{p}_{q}))$.

Providing excellent solutions often requires creative and practical thinking, where humans have a significant advantage over MLLMs: for a given question, humans can offer high-quality, diverse solutions, while MLLMs often struggle to generate ideas that are both innovative and feasible \cite{hubert2024current,jones2024bigger}. To help MLLMs balance answer quality and diversity during the optimization process, we formalize this task as a MOO problem, acknowledging its inherent multi-objective nature:
\begin{equation}
\text{Maximize}_{A\in\mathcal{A}}\mathcal {M}(A)=\left[\mathcal{M}^Q(A), \mathcal{M}^N(A)\right]^\top,
\label{eq1}
\end{equation}
where $\mathcal{A}=\mathcal{S}(f,\boldsymbol{p}_q)$ is the set of answers returned after the search strategy $\mathcal{S}$ is executed. $\mathcal{M}^Q(A)$ denotes the quality score of an answer, with higher values indicating a greater likelihood of correctness; $\mathcal{M}^N(A)$ represents the novelty score of answer $A$, with a higher score signifying greater distinctiveness compared to other answers. By optimizing for both $\mathcal{M}^Q(A)$ and $\mathcal{M}^N(A)$, the search process can identify a set of candidate answers that balance quality and diversity. Next, we delve into the design and calculation methods for the two scoring systems in detail.

\noindent\textbf{The formulation of $\mathcal{M}^Q(A)$.} An objective and fair quality scoring mechanism is crucial for enhancing the stability of $\mathcal{M}^Q(A)$. The existing scoring approach directly assigns a score to the generated answer $A$ using the MLLM $f$: $\mathcal{M}^Q(A) = f(\boldsymbol{p}_q, \boldsymbol{p}_s, A)$, where $\boldsymbol{p}_s$ is the scoring prompt \cite{BestaBKGPGGLNNH24,abs-2406-07394}. However, due to the sampling randomness in the MLLM generation process \cite{abs-2307-10236,abs-2401-12794}, this scoring mechanism may lead to inconsistencies and subjectivity, affecting the reliability of the quality scoring system. To address this, we introduce a reference-based scoring mechanism. For each question, we first use the MLLM to generate a reference answer $A_{\text{ref}}$, and then perform reference-based scoring in subsequent evaluations: $\mathcal{M}^Q(A) = f(\boldsymbol{p}_q, \boldsymbol{p}_s, A_{\text{ref}}, A)$, which provides a standard for the scoring process and reduces subjective bias in the MLLM scoring.

\noindent\textbf{The formulation of $\mathcal{M}^N(A)$.} Solving complex reasoning problems often requires more divergent thinking. Therefore, an effective novelty assessment mechanism is needed to facilitate the search for highly novel answers. A highly novel answer should exhibit significant differences from other answers, both in terms of surface form and semantic meaning. Therefore, we design the following evaluation metric, $\mathcal{M}^N(A)$, to assess an answer’s novelty based on semantic and textual uniqueness:
\begin{equation}
    \mathcal{M}^N(A) = \frac{D^{\text{Ed}}(A,\mathcal{A}^{/A})}{\max_{A\in\mathcal{A}}D^{\text{Ed}}(A,\mathcal{A}^{/A})} + D^{\text{SE}}(A,\mathcal{A}^{/A}),
    \label{eq2}
\end{equation}
where $D^{\text{Ed}}$ denotes the edit distance, and $\mathcal{A}^{/A}$ represents the set $\mathcal{A}$ excluding answer $A$. The term $D^{\text{Ed}}(A, \mathcal{A}^{/A}) = \sum_{A' \in \mathcal{A}^{/A}} D^{\text{Ed}}(A, A')$ refers to the sum of the edit distances between answer $A$ and the other answers in $\mathcal{A}$. We normalize the edit distance to the range $[0,1]$ using the maximum value of $D^{\text{Ed}}(A, \mathcal{A}^{/A})$ across all answers. And $D^{\text{SE}}(A,\mathcal{A}^{/A})$ is defined as
\begin{equation}
D^{\text{SE}}(A,\mathcal{A}^{/A})=\frac{1}{2\left|\mathcal{A}^{/A}\right|}\sum_{A\in\mathcal{A}^{/A}}\left(1-\frac{\langle E(A),E(A')\rangle}{\|E(A)\|\|E(A')\|}\right), 
\end{equation}
where $E$ represents the semantic encoder, which is used to extract semantic features for calculating the semantic similarity between answers. We instantiate $E$ using Sentence BERT \cite{ReimersG19}. $\left|\mathcal{A}^{/A}\right|$ denotes the number of elements in the set $\mathcal{A}^{/A}$. The inclusion of a factor of 2 in the denominator normalizes the range of $D^{\text{SE}}(A, \mathcal{A}^{/A})$ to $[0, 1]$, thereby eliminating the dimensionality discrepancy between $D^{\text{Ed}}$ and $D^{\text{SE}}$. Therefore, $D^{\text{SE}}$ represents the semantic distance between answer $A$ and the other answers in $\mathcal{A}^{/A}$.

With the MOO objective defined in equation \ref{eq1} and the corresponding quality and novelty evaluation metrics, we now introduce the EoT framework.

\subsection{Multi Objective Evolutionary Optimization}

Compared to single-objective optimization, multi-objective optimization (MOO) problems involve a larger and more complex solution space. Directly combining objectives into a single one can obscure important trade-offs and conflicts \cite{Zhou0Z024}, and determining appropriate weights is challenging. Therefore, an approach that effectively balances multiple independent objectives during optimization is crucial. Evolutionary algorithms, as global heuristic search methods, offer a more effective and comprehensive solution to MOO problems. Among these, NSGA-II is particularly well-suited due to its ability to efficiently handle multiple conflicting objectives while maintaining diversity in the solution set. Inspired by NSGA-II, the widely adopted MOO algorithm in evolutionary computation, we frame the optimization of MLLM reasoning as a genetic evolution process for answers. This approach strikes a balance between answer quality and novelty. The following sections describe each step in detail.

\noindent\textbf{Initialization. }
For each input question $\boldsymbol{p}_q$, we first use the MLLM to perform reasoning and generate $N$ initial candidate answers: $A_i = f(\boldsymbol{p}_q), i \in \{1, 2, \dots, N\}$, along with a reference answer $A_{\text{ref}}$. The candidate answers are then added to the candidate answer set $\mathcal{A}$: $\mathcal{A} = \{A_i\}_{i=1}^N$.

\noindent\textbf{Step 1: Scoring.}
For each candidate answer $A_i$ in the candidate answer set $\mathcal{A}$, calculate the quality score $\mathcal{M}^Q(A_i)$ and the novelty score $\mathcal{M}^N(A_i)$. Additionally, since adding new candidate answers affects the novelty scores of previously scored answers, we recalculate and update the novelty scores $\mathcal{M}^N(\tilde{A}_i)$ for all previously scored answers $\tilde{A}_i$ in each scoring round.

\noindent\textbf{Step 2: Ranking $\&$ Selection.}
\label{rank}
Since quality and novelty are often conflicting objectives \cite{le2024exploring} , a ranking algorithm is needed that balances both without prioritizing one over the other. To achieve this, we introduce the non-dominated sorting \cite{deb2002fast}, which concentrates high-quality candidates on the Pareto front, balancing the trade-offs between objectives. Specifically, for any two candidate answers $A_i$ and $A_j$, $A_i$ is considered dominated by $A_j$ ($A_i \prec A_j$) if and only if both $M_N(A_i) \leq M_N(A_j)$ and $M_Q(A_i) \leq M_Q(A_j)$ hold simultaneously. Conversely, $A_j$ dominates $A_i$ under the same condition. We calculate the dominance relationships for all pairs of candidates and rank them according to the number of candidates they dominate. Candidates that dominate more others are assigned higher priority levels (i.e., higher non-dominated levels), while those with the same dominance count are placed in the same level. This results in $L$ non-dominated levels, with each level $1 \leq l \leq L$ containing a set $\mathcal{A}_l$. Within each level, candidates are ranked according to their quality score $M_Q(A)$. Furthermore, we select $K$ parent candidates for the next generation by performing non-dominated sorting from the highest (starting at level 1) to the lowest levels. Within each non-dominated level, candidates are sorted by their quality score $M_Q(A)$. These parent candidates will then be used to generate offspring answers.

\noindent\textbf{Step 3: Crossover $\&$ Mutation.}
To guide the search process toward high-quality and diverse answers, we introduce crossover and mutation operations to continually optimize the existing candidates in terms of both quality and novelty, while generating offspring candidates. The crossover operation randomly selects two parent candidates $A'_i$ and $A'_j$ from the chosen parent candidate set $\{A'_i\}_{i=1}^K$, and summarizes them using the MLLM to produce a new high-quality offspring $A^{C}$ that retains the advantages of both: $A^{C}=f(\boldsymbol{p}_q,\boldsymbol{p}_C,A'_i,A'_j)$, where $\boldsymbol{p}_C$ is the crossover prompt. In the mutation operation, one parent candidate $A'_i$ is randomly selected, and the MLLM generates a novel offspring $A^M$ that is distinct from it, ensuring higher novelty:  
$ A^M = f(\boldsymbol{p}_q, \boldsymbol{p}_M, A'_i), $  where $ \boldsymbol{p}_M$ is the mutation prompt. By combining the designed crossover and mutation operations, we generate $N_C$ crossover offspring $\{A^C_{i}\}_{i=1}^{N_C}$ to search for high-quality answers, and simultaneously generate $N_M$ mutation offspring $\{A^M_{i}\}_{i=1}^{N_M}$ from $K$ parents to search for answers with high novelty, maintaining the ratio $r=\frac{N_C}{N_M}$ ($N_C+N_M=N$). Subsequently, we integrate the new offspring into the existing set of candidate answers: $\mathcal{A}=\mathcal{A}\cup\{A^C_{i}\}_{i=1}^{N_C}\cup \{A^M_{i}\}_{i=1}^{N_M}$ to update the candidate set and expand the number of offspring to $N\times=2$.

Throughout the optimization process, we conduct $T$ generations based on Steps 1-3 to obtain the final set of candidate answers. Since the offspring generation process is independent of any specific optimization path, it is fully parallelizable, thereby enhancing search efficiency compared to tree search algorithms. The pseudocode for the three steps of the EoT reasoning enhancement framework is presented in Appendix \ref{EoT_Algo}.

\subsection{Condensation $\&$ Aggregation Mechanism}
In practical MLLM-based question-answering systems, users typically expect a single, definitive answer. Simply returning a response based on the candidate set $ \mathcal{A} $ obtained through the MOO process, such as the Pass@k setting, is often insufficient to meet user needs. Therefore, it is necessary to enable the MLLM to summarize a precise answer from $ \mathcal{A} $. However, directly summarizing the entire set of candidates could lead to excessively long contexts, thereby constraining reasoning efficiency and accuracy, and potentially introducing low-quality answers that cause interference. Therefore, directly summarizing all candidate answers with the MLLM does not support accurate information aggregation. We propose the CA mechanism, which consists of two stages: condensation and aggregation. These stages extract the most valuable information from the candidate set $ \mathcal{A} $ to produce a precise and concise answer.

Specifically, in the condensation stage, we measure the distance $ D $ between different candidate answers $ A $ and $ A' $ in $\mathcal{A}$, based on their edit distance $ D^{\text{Ed}}(A, A') $ and semantic distance $ D^{SE}(A,A')=\frac{1}{2}-\frac{\langle E(A),E(A')\rangle}{2\|E(A)\|\|E(A')\|} $ (the scaling here is also for standardizing the magnitudes of $D^{\text{Ed}}$ and $D^{\text{SE}}$). The overall distance is computed as $ D(A, A')=\frac{D^{\text{Ed}}(A,A')}{\max_{A_i,A_j\in\mathcal{A}}D^{\text{Ed}}}+D^{\text{SE}}(A,A') $, yielding the distance matrix $\mathbf{D}(\mathcal{A},\mathcal{A})\in\mathbb R^{N\times  N}$, where $\mathbf{D}(\mathcal{A},\mathcal{A})_{i,j}=D(A_{i},A_{j})$.

Next, we apply K-Medoids clustering \cite{NewlingF17} on $ D(\mathcal{A}, \mathcal{A}) $ to form $ \kappa $ clusters $ \{\mathcal{C}_j\}_{j=1}^\kappa $, with the center of each cluster defined as the answer $ A^j_c = \arg \min_{A_c \in \mathcal{C}_j} \frac{1}{|\mathcal{C}_j|} \sum_{A' \in \mathcal{C}_j} D(A_c, A') $, which minimizes the average distance to all other answers in the cluster. We then compute the average quality score for each cluster:$
\bar{\mathcal{M}}^Q(\mathcal{C}_j) = \frac{1}{|\mathcal{C}_j|} \sum_{A \in \mathcal{C}_j} \mathcal{M}^Q(A)$, and sort the clusters in descending order based on $\bar{\mathcal{M}}^Q(\mathcal{C}_j)$. To eliminate the influence of low-quality answers, we remove the $ m $ clusters with the lowest average quality scores. From the remaining $ \kappa - m $ clusters, we select the center answer from each one to form the condensed candidate set $ \{A^j_c\}_{j=1}^{\kappa - m} $, which helps the MLLM more efficiently extract precise information from a large number of candidates. Finally, we pass the condensed candidate set to the MLLM for aggregation, resulting in a clear, accurate answer $ A^* = f(\boldsymbol{p}_q, \boldsymbol{p}_A, \{A^j_c\}_{j=1}^{\kappa - m}) $, where $ \boldsymbol{p}_A $ is the aggregation prompt.

In summary, EoT serves as a versatile and efficient reasoning enhancement framework that balances answer quality and diversity during the search process, while effectively generating accurate final answers through the CA mechanism as required.

%% file: sec/4_Experiments.tex
\section{Experiments}
\label{sec:Experiments}

\begin{table*}[htbp]
\renewcommand{\arraystretch}{0.85}
% \vspace{-5pt}
\centering
\resizebox{0.85\textwidth}{!}{% 将表格宽度调整为页面的 80%
\begin{tabular}{ccclcccc}
\toprule[1.5pt]
\rowcolor{gray!50}
\textbf{Model} & \textbf{Size} & \textbf{Open Source}& \textbf{Method}  & \textbf{MathVista} & \textbf{Math-Vision} & \textbf{GSM8K} & \textbf{AVG}  \\
\midrule
Gemini 1.5 Pro & - & \ding{56} & \cellcolor{gray!10}  CoT   & \cellcolor{gray!10} 52.10 & \cellcolor{gray!10} 17.11 & \cellcolor{gray!10} 86.50  & \cellcolor{gray!10} 51.90 \\
GPT-4V & - & \ding{56} & CoT & 49.90 & 22.37 & 87.10 & 53.12 \\
\midrule
\multirow{8}{*}{\rotatebox{45}{\large{Qwen2-VL}}}
& \multirow{8}{*}{7B} & \multirow{8}{*}{\ding{52}}
        & \cellcolor{gray!10}  IO & \cellcolor{gray!10} $33.25$ & \cellcolor{gray!10} 10.20 & \cellcolor{gray!10} 50.00 & \cellcolor{gray!10} 31.15\\
&   &   & CoT & ${38.75}_{\textcolor{red}{\uparrow5.50}}$ & $16.78_{\textcolor{red}{\uparrow6.58}}$ & $66.25_{\textcolor{red}{\uparrow16.25}}$ & $40.59_{\textcolor{red}{\uparrow9.44}}$ \\
&   &   & \cellcolor{gray!10}  Self-Refine & \cellcolor{gray!10} $37.75_{\textcolor{red}{\uparrow4.50}}$ & \cellcolor{gray!10} $16.78_{\textcolor{red}{\uparrow6.58}}$ & \cellcolor{gray!10} $68.50_{\textcolor{red}{\uparrow18.50}}$ & \cellcolor{gray!10} $41.01_{\textcolor{red}{\uparrow9.86}}$ \\
&   &   &MCTSr & $31.25_{\textcolor{blue}{\downarrow2.00}}$ & $15.46_{\textcolor{red}{\uparrow5.26}}$ & $70.00_{\textcolor{red}{\uparrow20.00}}$ & $38.90_{\textcolor{red}{\uparrow7.75}}$ \\
&   &   & \cellcolor{gray!10}  SPP & \cellcolor{gray!10} $31.00_{\textcolor{blue}{\downarrow2.25}}$ & \cellcolor{gray!10} $9.21_{\textcolor{blue}{\downarrow0.99}}$ & \cellcolor{gray!10} $62.75_{\textcolor{red}{\uparrow12.75}}$ & \cellcolor{gray!10} $34.32_{\textcolor{red}{\uparrow3.17}}$ \\
&   &   &MAD & $45.00_{\textcolor{red}{\uparrow11.75}}$ & $12.83_{\textcolor{red}{\uparrow2.63}}$ & $74.00_{\textcolor{red}{\uparrow24.00}}$ & $43.94_{\textcolor{red}{\uparrow12.79}}$ \\
&   &   & \cellcolor{gray!10}  ToT & \cellcolor{gray!10} $38.25_{\textcolor{red}{\uparrow5.00}}$ & \cellcolor{gray!10} $15.79_{\textcolor{red}{\uparrow5.59}}$ & \cellcolor{gray!10} $66.75_{\textcolor{red}{\uparrow16.75}}$ & \cellcolor{gray!10} $40.26_{\textcolor{red}{\uparrow9.11}}$ \\
&   &   & EoT & $\textbf{45.42}_{\textcolor{red}{\uparrow12.17}}$ & $\textbf{17.11}_{\textcolor{red}{\uparrow6.91}}$  & $\textbf{76.50}_{\textcolor{red}{\uparrow26.50}}$  & $\textbf{46.20}_{\textcolor{red}{\uparrow15.05}}$ \\
\midrule
\multirow{8}{*}{\rotatebox{45}{\large{LLaVA-NeXT}}}
& \multirow{8}{*}{8B} & \multirow{8}{*}{\ding{52}}
        & \cellcolor{gray!10}  IO & \cellcolor{gray!10} 16.50 & \cellcolor{gray!10} 11.84 & \cellcolor{gray!10} 7.25 & \cellcolor{gray!10} 11.86 \\
&   &   &CoT & $21.00_{\textcolor{red}{\uparrow4.5}}$ & $12.17_{\textcolor{red}{\uparrow0.33}}$ & $35.25_{\textcolor{red}{\uparrow5.26}}$ & $22.81_{\textcolor{red}{\uparrow10.95}}$ \\
&   &   & \cellcolor{gray!10}  Self-Refine & \cellcolor{gray!10}$20.50_{\textcolor{red}{\uparrow4.00}}$ & \cellcolor{gray!10} $12.54_{\textcolor{red}{\uparrow0.70}}$ & \cellcolor{gray!10} $32.50_{\textcolor{red}{\uparrow25.25}}$ & \cellcolor{gray!10} $21.85_{\textcolor{red}{\uparrow9.99}}$ \\
&   &   &MCTSr & $21.00_{\textcolor{red}{\uparrow4.50}}$ & $11.55_{\textcolor{blue}{\downarrow0.29}}$ & $36.25_{\textcolor{red}{\uparrow 29.00}}$ & $22.93_{\textcolor{red}{\uparrow 11.07}}$ \\
&   &   & \cellcolor{gray!10}  SPP & \cellcolor{gray!10} $15.50_{\textcolor{blue}{\downarrow 1.00}}$ & \cellcolor{gray!10} $8.88_{\textcolor{blue}{\downarrow 2.96}}$  & \cellcolor{gray!10} $30.00_{\textcolor{red}{\uparrow22.75}}$ & \cellcolor{gray!10} $18.13_{\textcolor{red}{\uparrow6.27}}$ \\
&   &   &MAD & $21.00_{\textcolor{red}{\uparrow4.50}}$ & $12.50_{\textcolor{red}{\uparrow0.66}}$ & $37.75_{\textcolor{red}{\uparrow30.50}}$ & $23.75_{\textcolor{red}{\uparrow11.89}}$ \\
&   &   & \cellcolor{gray!10}  ToT & \cellcolor{gray!10} $20.25_{\textcolor{red}{\uparrow3.75}}$ & \cellcolor{gray!10} $9.54_{\textcolor{blue}{\downarrow2.30}}$  & \cellcolor{gray!10} $46.00_{\textcolor{red}{\uparrow38.75}}$ & \cellcolor{gray!10} $25.26_{\textcolor{red}{\uparrow13.40}}$ \\
&   &   &EoT & $\textbf{21.50}_{\textcolor{red}{\uparrow5.00}}$ & $\textbf{13.49}_{\textcolor{red}{\uparrow1.65}}$ & $\textbf{49.00}_{\textcolor{red}{\uparrow42.75}}$ & $\textbf{28.33}_{\textcolor{red}{\uparrow16.47}}$ \\
\midrule
\multirow{8}{*}{\rotatebox{45}{\large{Phi-3.5-vision}}}
& \multirow{8}{*}{4.5B} & \multirow{8}{*}{\ding{52}}
        & \cellcolor{gray!10}  IO & \cellcolor{gray!10} 25.75 & \cellcolor{gray!10} 7.89 &\cellcolor{gray!10}  11.25 & \cellcolor{gray!10} 14.96 \\
&   &   &CoT & $27.00_{\textcolor{red}{\uparrow1.25}}$ & $12.83_{\textcolor{red}{\uparrow4.94}}$ & $53.75_{\textcolor{red}{\uparrow42.20}}$ & $31.20_{\textcolor{red}{\uparrow16.24}}$ \\
&   &   & \cellcolor{gray!10}  Self-Refine & \cellcolor{gray!10} $24.00_{\textcolor{blue}{\downarrow1.75}}$ & \cellcolor{gray!10} $10.86_{\textcolor{red}{\uparrow2.97}}$ & \cellcolor{gray!10} $48.25_{\textcolor{red}{\uparrow37.00}}$ & \cellcolor{gray!10} $27.70_{\textcolor{red}{\uparrow12.74}}$ \\
&   &   &MCTSr& $20.50_{\textcolor{blue}{\downarrow5.25}}$ & $11.18_{\textcolor{red}{\uparrow3.29}}$ & $39.25_{\textcolor{red}{\uparrow29.00}}$ & $23.64_{\textcolor{red}{\uparrow8.68}}$ \\
&   &   & \cellcolor{gray!10}  SPP & \cellcolor{gray!10} $13.25_{\textcolor{blue}{\downarrow12.50}}$ & \cellcolor{gray!10} $10.86_{\textcolor{red}{\uparrow2.97}}$ & \cellcolor{gray!10} $52.00_{\textcolor{red}{\uparrow40.75}}$ & \cellcolor{gray!10} $25.37_{\textcolor{red}{\uparrow10.41}}$ \\
&   &   &MAD & $29.50_{\textcolor{red}{\uparrow3.75}}$ & $12.28_{\textcolor{red}{\uparrow4.39}}$ & $61.50_{\textcolor{red}{\uparrow50.25}}$ & $34.43_{\textcolor{red}{\uparrow19.47}}$ \\
&   &   & \cellcolor{gray!10}  ToT & \cellcolor{gray!10} $20.75_{\textcolor{blue}{\downarrow5.00}}$ & \cellcolor{gray!10} $9.21_{\textcolor{red}{\uparrow1.32}}$  & \cellcolor{gray!10} $46.50_{\textcolor{red}{\uparrow35.25}}$ & \cellcolor{gray!10} $25.49_{\textcolor{red}{\uparrow10.53}}$ \\
&   &   &EoT & $\textbf{29.98}_{\textcolor{red}{\uparrow4.23}}$ & $\textbf{13.82}_{\textcolor{red}{\uparrow5.93}}$ & $\textbf{62.35}_{\textcolor{red}{\uparrow51.10}}$ & $\textbf{34.94}_{\textcolor{red}{\uparrow19.98}}$ \\

\bottomrule[1.5pt]
\end{tabular}
% \vspace{-5pt}
}

\caption{Performance (\%) of vision-language models using different reasoning methods on mathematical reasoning benchmarks (MathVista, Math-Vision, GSM8K) and their average accuracy (AVG).}
\label{pass@1}
\end{table*}

\begin{table*}[htbp]
\renewcommand{\arraystretch}{0.9}
% \vspace{-5pt}
\centering
\resizebox{0.85\textwidth}{!}{ 
\begin{tabular}{cccccccccc}
\toprule[1.5pt]
\multirow{2}{*}{\textbf{Model}} & \multirow{2}{*}{\textbf{Method}} &  \multicolumn{4}{c}{\textbf{Pass@4}} & \multicolumn{4}{c}{\textbf{Pass@8}} \\ 
&   & MathVista  &   Math-Vision  &  GSM8K  & AVG &  MathVista  &   Math-Vision  &  GSM8K & AVG \\ 
\midrule
\multirow{3}{*}{Qwen2-VL} 
&\cellcolor{gray!10} ToT   & \cellcolor{gray!10} 41.75 & \cellcolor{gray!10} 19.08 & \cellcolor{gray!10} 72.25 & \cellcolor{gray!10} 44.36 & \cellcolor{gray!10} 55.00 & \cellcolor{gray!10} 37.50 & \cellcolor{gray!10} 83.50 & \cellcolor{gray!10} 58.67 \\
& MCTSr & 56.50 & 34.21 & 88.00 & 59.57 & 66.75 & 50.66 & 93.00 & 70.14 \\
&\cellcolor{gray!10}  EoT   & \cellcolor{gray!10} \textbf{58.75} & \cellcolor{gray!10} \textbf{34.87} & \cellcolor{gray!10} \textbf{90.25} & \cellcolor{gray!10} \textbf{61.29} & \cellcolor{gray!10} \textbf{67.50} & \cellcolor{gray!10} \textbf{51.64} & \cellcolor{gray!10} \textbf{94.25} & \cellcolor{gray!10} \textbf{71.13} \\
\midrule
\multirow{3}{*}{LLaVA-NeXT} 
& \cellcolor{gray!10} ToT   & \cellcolor{gray!10} 24.00 & \cellcolor{gray!10} 14.80 & \cellcolor{gray!10} 49.00 & \cellcolor{gray!10} 29.27 & \cellcolor{gray!10} 55.00 & \cellcolor{gray!10} 37.50 & \cellcolor{gray!10} 63.50 & \cellcolor{gray!10} 52.00 \\
& MCTSr & 37.75 & 28.95 & 66.75 & 44.48 & \textbf{48.75} & 51.64 & 78.50 & 59.63 \\
& \cellcolor{gray!10}  EoT   & \cellcolor{gray!10} \textbf{38.75} & \cellcolor{gray!10} \textbf{37.50} & \cellcolor{gray!10} \textbf{68.50} & \cellcolor{gray!10} \textbf{48.25} & \cellcolor{gray!10} 48.50 & \cellcolor{gray!10} \textbf{52.33} & \cellcolor{gray!10} \textbf{79.25} & \cellcolor{gray!10} \textbf{61.73} \\
\midrule
\multirow{3}{*}{Phi-3.5-vision} 
& \cellcolor{gray!10}  ToT   & \cellcolor{gray!10} 29.25 & \cellcolor{gray!10} 15.79 & \cellcolor{gray!10} 56.75 & \cellcolor{gray!10} 33.93 & \cellcolor{gray!10} 40.25 & \cellcolor{gray!10} 31.25 & \cellcolor{gray!10} 78.00 & \cellcolor{gray!10} 49.83 \\
& MCTSr & 39.00 & 27.30 & 76.00 & 47.43 & 49.25 & 41.12 & 82.75 & 57.71 \\
& \cellcolor{gray!10}  EoT   & \cellcolor{gray!10} \textbf{43.00} & \cellcolor{gray!10} \textbf{37.17} & \cellcolor{gray!10} \textbf{85.75} & \cellcolor{gray!10} \textbf{55.31} & \cellcolor{gray!10} \textbf{55.75} & \cellcolor{gray!10} \textbf{50.99} & \cellcolor{gray!10} \textbf{90.25} & \cellcolor{gray!10} \textbf{65.66} \\ 

\bottomrule[1.5pt]
\end{tabular}
}
\caption{Pass@4 and Pass@8 performance (\%) of vision-language models using reasoning methods (ToT, MCTSr, EoT) on mathematical reasoning benchmarks (MathVista, Math-Vision, GSM8K) and their average accuracy (AVG)}
\label{pass@k}
\vspace{-10pt}
\end{table*}

 \textbf{Models and Datasets. }To evaluate the effectiveness of our approach, we employ three based models: Qwen2VL \cite{Qwen2VL}, Llava \cite{liu2024llavanext}, and Phi-3.5 \cite{microsoft2024phi35visioninstruct}. These three models are all common multimodal models. For benchmarking, we use test sets from MathVista \cite{lu2024mathvista}, Math-Vision \cite{wang2024MATHVision}, and GSM8K \cite{cobbe2021gsm8k}, where MathVista and Math-Vision are multimodal datasets and GSM8K is a text-only dataset. To evaluate answer accuracy, we instruct the models to respond in the format “\verb|The Answer is \boxed{{}}|” and then extract the final answers for comparison against ground truth labels. An answer is considered correct if it matches the ground truth exactly. All experiments were conducted on a single Nvidia A100 80GB GPU.

\noindent \textbf{Baselines. }

We compare EoT with several existing methods, including closed-source models such as Gemini \cite{team2023gemini} and GPT-4V \cite{yang2023dawn}, as well as open methods like IO, CoT \cite{Wei0SBIXCLZ22}, Self-Refine \cite{Self-refine}, MCTSr \cite{abs-2406-07394}, SPP \cite{wang2023unleashing}, MAD \cite{liang2023encouraging}, and ToT \cite{YaoYZS00N23}. For the IO baseline, we use a standard zero-shot input-output prompt. In CoT, we apply prompting formats such as “[Reasoning process]…[Verification]” and “Let’s think step by step” to encourage the model to generate reasoning steps. Self-Refine is implemented with one refinement iteration, while MCTSr uses four expansion steps. SPP is implemented using the standard prompt defined in \cite{wang2023unleashing}. For MAD, we set the number of agents to 3 and conduct two rounds of discussion. ToT utilize a greedy search strategy with two rounds of exploration, sampling five answers per round and selecting the best answer via voting for further reasoning. 
Detailed prompts for EoT can be found in the Appendix \ref{prompts}.

\begin{figure*}[t]
% \vspace{-5pt}
	\centering
\includegraphics[width=1\textwidth]{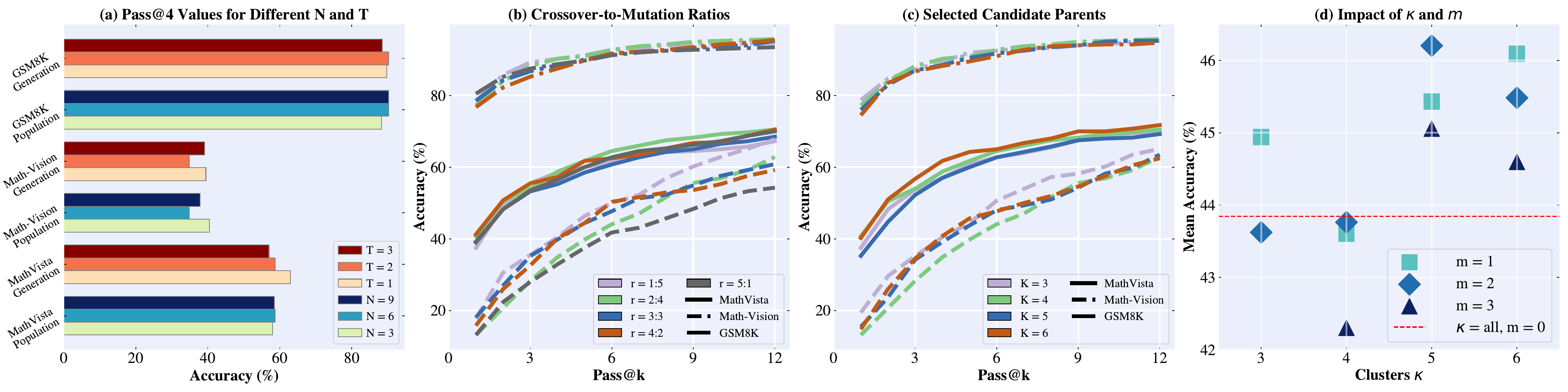}
% \vspace{-10pt}
	\caption{Abalation results on different $N,T,r,K,\kappa,m$ parameters and datsets. Fig (a) shows pass@4 accuracy (\%) on different candidate answers $N$ and evolutionary generations $T$. Fig (b) and (c) show the pass@$k$ ($k=1,2,...,12$) accuracy (\%) curves under different crossover-to-mutation ratios $r$ and selected candidate parents $K$. Fig (d) shows the average accuracy on three datasets under different clusters $\kappa$ and drop clusters $m$.}
	\label{abalation} 
    % \vspace{-10pt}
\end{figure*}

\subsection{Evaluation of Diversity}
The Pass@K metric is an evaluation metric used to measure the accuracy of MLLM in generating answers, specifically focusing on whether the correct answer appears within the top K generated answers. More precisely, Pass@K is defined as the probability that the correct answer is among the top K answers generated by the model. It can be calculated as Pass@$k$:$ = \frac{1}{N} \sum_{N} \mathbbm{1} \left(G \in \mathcal{A}_k \right)$, where 
$N$ is the total number of test cases, $\mathcal{A}_k$:$=\{A_i\}_{i=1}^k$ denotes the set of candidate answers, and $\mathbb{I}$ is the indicator function. This metric provides a comprehensive evaluation of the model's performance for a given K value.
To compare the performance of different models based on the Pass@K metric, we conduct a comparative analysis of methods that generate answer lists during the answering process, such as EoT, MCTSr, and ToT. For MCTSr, the answer list is generated by the model during the exploration process and sorted by answer scores from high to low. For ToT, the answer list is sorted by rounds in reverse order, with answers within the same round sorted by vote count from high to low. For EoT, the answer list is sorted based on non-dominated sorting. The experimental results are detailed in Table \ref{pass@k}.

Based on the experimental results, under the Pass@$k$ evaluation metric, applying EoT to the three models achieved superior performance on the MathVista, Math-Vision, and GSM8K datasets compared to ToT and MCTSr. This demonstrates that EoT not only maintains answer diversity but also improves accuracy in generating correct answers. These results highlight that EoT, while preserving diversity, delivers higher-quality answer generation than alternative methods.

\subsection{Evaluation of Quality}
To assess the impact of introducing the diversity metric $\mathcal{M}^{N}(A)$ on answer quality during the reasoning path search, we evaluate EoT using Pass@$1$ on the MathVista, Math-Vision, and GSM8K datasets. The results, present in Table \ref{pass@1}, indicate that EoT outperforms other methods across all datasets when using the same base models. This suggests that incorporating the diversity metric positively enhances the model’s reasoning capabilities. While EoT shows slightly lower performance than Gemini 1.5 Pro and GPT-4V, this may be due to the significantly larger parameter sizes of Gemini 1.5 Pro and GPT-4V, which provide substantially stronger reasoning abilities compared to our chosen base models.

\subsection{Evaluation of  Efficiency}
Additionally, to demonstrate the efficiency of the EoT method, we compare EoT with ToT and MCTSr in terms of the number of reasoning steps required to generate $K$ answers, as well as the average time taken to generate each answer when producing 12 answers. The experimental results are presented in Table \ref{efficiency}. From the results, it can be observed that, under the condition of generating the same number of $K$ answers, EoT requires fewer reasoning steps and less time per step. This indicates that the EoT method achieves higher reasoning efficiency and lower overhead compared to other methods.

\subsection{Ablation Studies}
 To better demonstrate the excellent properties of EoT, we design four sets of ablation experiments. Through these experiments, we aim to answer three key questions.

 \textbf{Q1: Does the advantage of EoT rely on a large population size and more evolutionary generations?} To address this question, we evaluate the model's pass@4 metric under different candidate answers $N$ and the number of generations $T$, specifically with $N = 3, 6, 9$ and $T = 1, 2, 3$. The experimental results are presented in Figure \ref{abalation} (a). From the experimental results, we observe that EoT exhibits highly consistent pass@4 inference performance across different values of $N$ and $T$, with the maximum performance fluctuation being just 5.13\%. This indicates that the outstanding performance of EoT does not depend on generating a large number of mutated offspring, meaning that EoT can robustly enhance the model’s inference capability without significantly increasing computational resource consumption.

 \textbf{Q2: How do the number of candidate parent nodes and the crossover/mutation ratio impact model inference?} We set the crossover-to-mutation ratios to $1:1, 1:2, 1:5, 5:1$, and $2:1$; and selected candidate parents $K=3,4,5,6$ from non-dominated sorted set $\mathcal{A}_l$. The experimental results are shown in Figure \ref{abalation} (b,c). From the results, we observe that on the simpler Math-Vista and GSM8K datasets, the pass@$k$ curves exhibit high consistency across different settings. On the more challenging Math-Vision dataset, the configuration with $r=1:5$ and $K=3$ demonstrates the best performance. This suggests that, in complex reasoning scenarios, a more diverse offspring population and high-quality parents can significantly enhance the model's reasoning performance.

 \textbf{Q3: How does the CA mechanism impact the model's performance when generating diverse answers?} To address this question, we divide the generated answers into $3, 4, 5$, and $6$ clusters, and drop the $1, 2$, and $3$ clusters with the poorest quality answers. We then compare how the model's average performance changes across three datasets as the number of clusters increases, reflecting the increase in answer diversity. The experimental results are shown in Figure \ref{abalation} (d). From this, we can observe that as the diversity of generated answers increases, the CA mechanism can effectively help improve model performance. However, when the number of clusters is $\kappa \leq 4$, the CA mechanism does not help performance improvements. This may be because the quality of the generated answers is already high, and drop certain answers limits the model's potential.

\begin{table}[ht]
\centering
\begin{tabular}{ccc}
\toprule
Method & Inference Steps & \parbox{2.5cm}{\centering Time per Answer\\($K=12$)} \\
\midrule
ToT    & $2K$     & $14.62$ \\
MCTSr  & $3K - 1$ & $26.36$ \\
EoT    & $2K + 1$ & $15.23$ \\
\bottomrule
\end{tabular}
\caption{Inference Efficiency for $K$ Answers (time: \textit{s})}
\label{efficiency}
\vspace{-15pt}
\end{table}

%% file: sec/5_Conclusion.tex
\section{Conclusion}

In this work, we propose the EOT framework, which formulates the search for reasoning paths as a MOO problem, balancing both quality and diversity.
EOT overcomes the limitations of local optima in traditional methods by employing non-dominated sorting to explore candidate solutions along the Pareto front, achieving a balanced improvement in both quality and diversity.
By utilizing specially designed crossover and mutation operations, EOT enables offspring solutions to inherit strengths from diverse parent solutions, consistently producing diverse reasoning results.
EOT further incorporates an efficient CA mechanism that streamlines candidate answers based on semantic similarity, preserving both diversity and quality to enable accurate and efficient answer summarization for MLLMs.
We believe this work provides a valuable new perspective for designing advanced reasoning frameworks based on heuristic methods.

%% file: sec/X_suppl.tex
\clearpage
\setcounter{page}{1}
\maketitlesupplementary

\appendix

\section{Algorithm of EoT}
Here, we introduce the algorithmic description of EoT, with its workflow depicted in Algorithm \ref{EoT_Algo}.
\label{sec:Algo}
\begin{algorithm}[h]
\caption{EoT Search Algorithm}
\label{EoT_Algo}
\begin{algorithmic}[1]
\STATE \textbf{Input:} User query prompt $\bm{p}_q$, LLM function $f$, evaluation metric $\mathcal{M}$, number of generations $T$, number of candidates $N$, crossover rate $r$
\STATE \textbf{Output:} Optimized set of candidate answers $\mathcal{A}$

\STATE Initialize candidate set $\mathcal{A} = \{ A_i \}_{i=1}^N$, where each $A_i = f(\bm{p}_q)$ is generated by the LLM
\STATE Generate a reference answer $A_{\text{ref}}$ using LLM
\STATE Calculate the initial quality scores $\mathcal{M}^Q(A_i)$ and novelty scores $\mathcal{M}^N(A_i)$ for each $A_i \in \mathcal{A}$

\FOR{each generation $t=1, 2, \dots, T$}
    \STATE \textbf{Step 1: Scoring}
    \FOR{each candidate answer $A_i \in \mathcal{A}$}
        \STATE Recalculate the quality score $\mathcal{M}^Q(A_i)$ based on reference-based scoring
        \STATE Recalculate the novelty score $\mathcal{M}^N(A_i)$ for all candidates in $\mathcal{A}$, including updates due to new candidates
    \ENDFOR

    \STATE \textbf{Step 2: Ranking and Selection}
    \STATE Perform non-dominated sorting on the candidates based on their $\mathcal{M}^Q$ and $\mathcal{M}^N$ scores
    \STATE Sort candidates into $L$ non-dominated levels based on dominance relations
    \STATE Select $K$ parent candidates from the highest-ranked non-dominated levels (sorted by $\mathcal{M}^Q$ within levels)

    \STATE \textbf{Step 3: Crossover and Mutation}
    \STATE Generate $N_C$ crossover offspring $\{A^C_i\}_{i=1}^{N_C}$ using selected parent pairs $A'_i$ and $A'_j$: 
    \STATE $A^C = f(\bm{p}_q, \bm{p}_C, A'_i, A'_j)$
    \STATE Generate $N_M$ mutation offspring $\{A^M_i\}_{i=1}^{N_M}$ using randomly selected parent $A'_i$: 
    \STATE $A^M = f(\bm{p}_q, \bm{p}_M, A'_i)$
    \STATE Update the candidate set $\mathcal{A} = \mathcal{A} \cup \{ A^C_i \}_{i=1}^{N_C} \cup \{ A^M_i \}_{i=1}^{N_M}$

    \STATE Expand the candidate set size: $N \times= 2$
\ENDFOR

\STATE \textbf{Return:} Optimized set of candidate answers $\mathcal{A}$

\end{algorithmic}
\end{algorithm}

\section{Configurations in Experiment}
%实验设置
\subsection{Experimental Environment}
Each experiment conducted on a high-performance server with the following hardware and software configurations:

\begin{itemize}
    \item \textbf{Hardware}: NVIDIA A100 GPU 80GB and an Intel(R) Xeon(R) Processor @ 2.90GHz.
    \item \textbf{Operating System}: Ubuntu 22.04 LTS.
    \item \textbf{Software}: Python 3.10.6, PyTorch 2.4.1, along with essential libraries such as NumPy 1.26.3 and Transformers 4.45.2.
\end{itemize}

\subsection{Model Configurations}
Our experiments utilize three state-of-the-art vision-language models for reasoning tasks:

\begin{enumerate}
    \item \textbf{Qwen2VL}: The latest multimodal model from the Qwen family. We use the 7B parameter version, released in August 2024.
    \item \textbf{LLaVA-NeXT}: We use the version based on LLaMA3-8B, released in May 2024. This model is trained with a more advanced language model, improving its performance and capabilities.
    \item \textbf{Phi3.5-vision}: A lightweight multimodal model developed by the Microsoft team with 4.5B parameters, released in August 2024.
\end{enumerate}
The model weights used in our work are sourced from Hugging Face, and we deploy the models using the Transformers library, utilizing Flash-Attention2 for inference acceleration.

\subsection{Comparative Methods}
As mentioned in \ref{sec:Experiments}, we evaluate our framework by comparing it with state-of-the-art methods:
\begin{enumerate}
    \item \textbf{IO (Input-Output)}: We use a standard zero-shot input-output prompt, where the model will only response with “\verb|The Answer is \boxed{{}}|”.
    \item \textbf{CoT (Chain-of-Thought)}: We apply prompting formats as “[Reasoning process]...[Verification]...” and “Let’s think step by step” to encourage the model to generate reasoning steps and validation steps.
    \item \textbf{Self-Refine}: Self-Refine improves model outputs through one refinement iteration, without requiring extra training or supervision. The process involves first providing feedback on the initial output, followed by refining the answer based on that feedback. We use the second response from MCTSr for this method.
    \item \textbf{ToT (Tree of Thoughts)}: ToT employs a greedy search strategy with two rounds of exploration. In each round, five answers are sampled, and the best answer is selected via voting for further reasoning. We adopt the original settings from \cite{YaoYZS00N23}.
    \item \textbf{MCTSr (Monte Carlo Tree Self-Refine)}: MCTSr combines Monte Carlo Tree Search with self-refine to enhance decision-making and solve complex reasoning tasks. We adopt the original settings from \cite{abs-2406-07394}.
    \item \textbf{SPP (Solo Performance Prompting)}: SPP enables a single model to collaborate with itself through multiple personas in multi-turn interactions. We adopt the original settings from \cite{wang2023unleashing}.
    \item \textbf{MAD (Multi-Agent Debate)}: MAD utilizes multiple agents engaged in debate, guided by a judge, to promote divergent thinking and address complex reasoning tasks. We adopt the original settings from \cite{liang2023encouraging}.
\end{enumerate}

All prompts are zero-shot, with the model required to end the response with “\verb|The Answer is \boxed{{}}|”. The output is extracted and compared to the true label. Throughout the shared steps, prompts are kept consistent across the different methods.

\section{Prompt Examples and Corresponding Answers in Experiments}
\label{prompts}
\subsection{User Query Prompt \texorpdfstring{$\bm{p}_q$}{$p_q$}}
\begin{figure}[h]
    \centering
    \vspace{-5mm}
    \includegraphics[width=\linewidth]{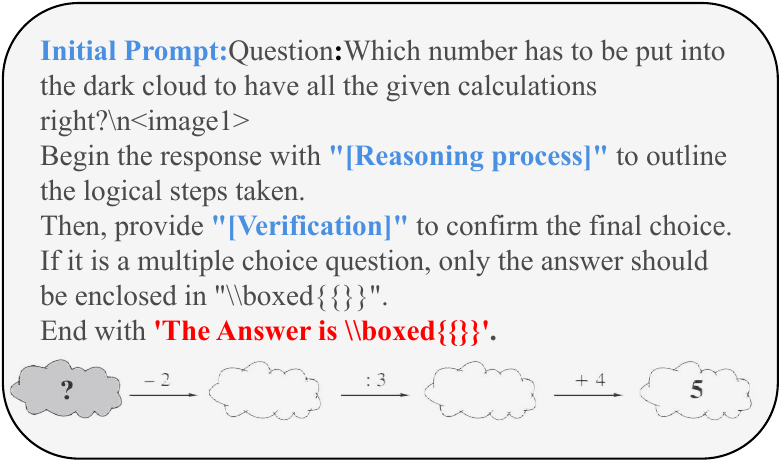}
    \label{prompts:initial_prompt}
    \vspace{-10mm}
\end{figure}

\subsection{Quality Scoring Prompt \texorpdfstring{$\bm{p}_s$}{$p_s$}}
\begin{figure}[H]
    \centering
    \vspace{-5mm}
    \includegraphics[width=\linewidth]{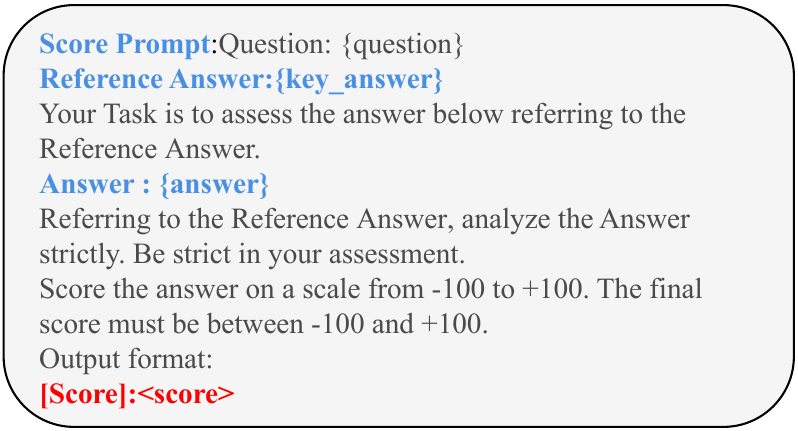}
    \label{prompts:score_prompt}
    \vspace{-8mm}
\end{figure}

\subsection{Crossover Prompt \texorpdfstring{$\bm{p}_C$}{$p_C$}}
\begin{figure}[H]
    \centering
    \vspace{-2mm}
    \includegraphics[width=\linewidth]{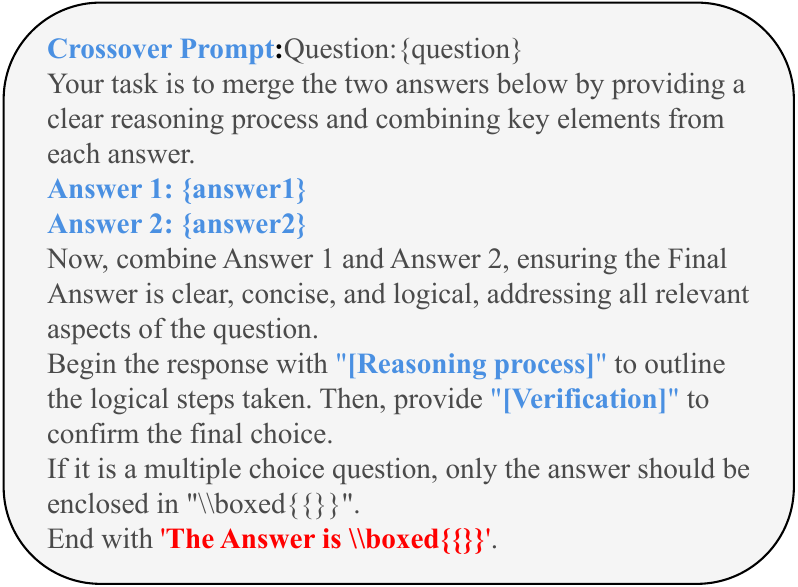}
    \label{prompts:cross_prompt}
    \vspace{-4mm}
\end{figure}

\subsection{Mutate Prompt \texorpdfstring{$\bm{p}_M$}{$p_M$}}
\begin{figure}[H]
    \centering
    \vspace{-2mm}
    \includegraphics[width=\linewidth]{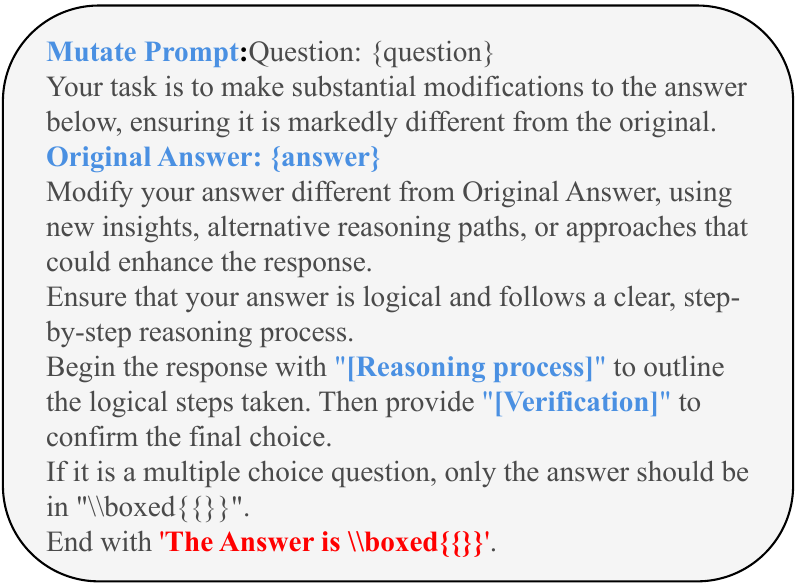}
    \label{prompts:mutate_prompt}
    \vspace{-4mm}
\end{figure}

\subsection{Aggregation Prompt \texorpdfstring{$\bm{p}_A$}{$p_A$}}
\begin{figure}[H]
    \centering
    \vspace{-2mm}
    \includegraphics[width=\linewidth]{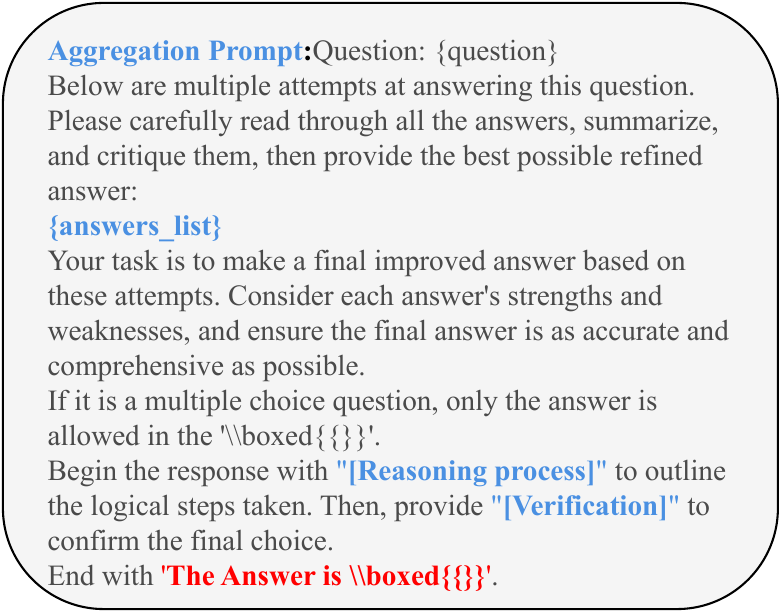}
    \label{prompts:aggregation_prompt}
    \vspace{-4mm}
\end{figure}

\newpage

\subsection{Candidate Answers Set \texorpdfstring{$\mathcal{A}$}{A}}
\begin{figure}[H]
    \centering
    \vspace{-1mm}
    \hspace*{-9mm}\includegraphics[width=1.18\linewidth]{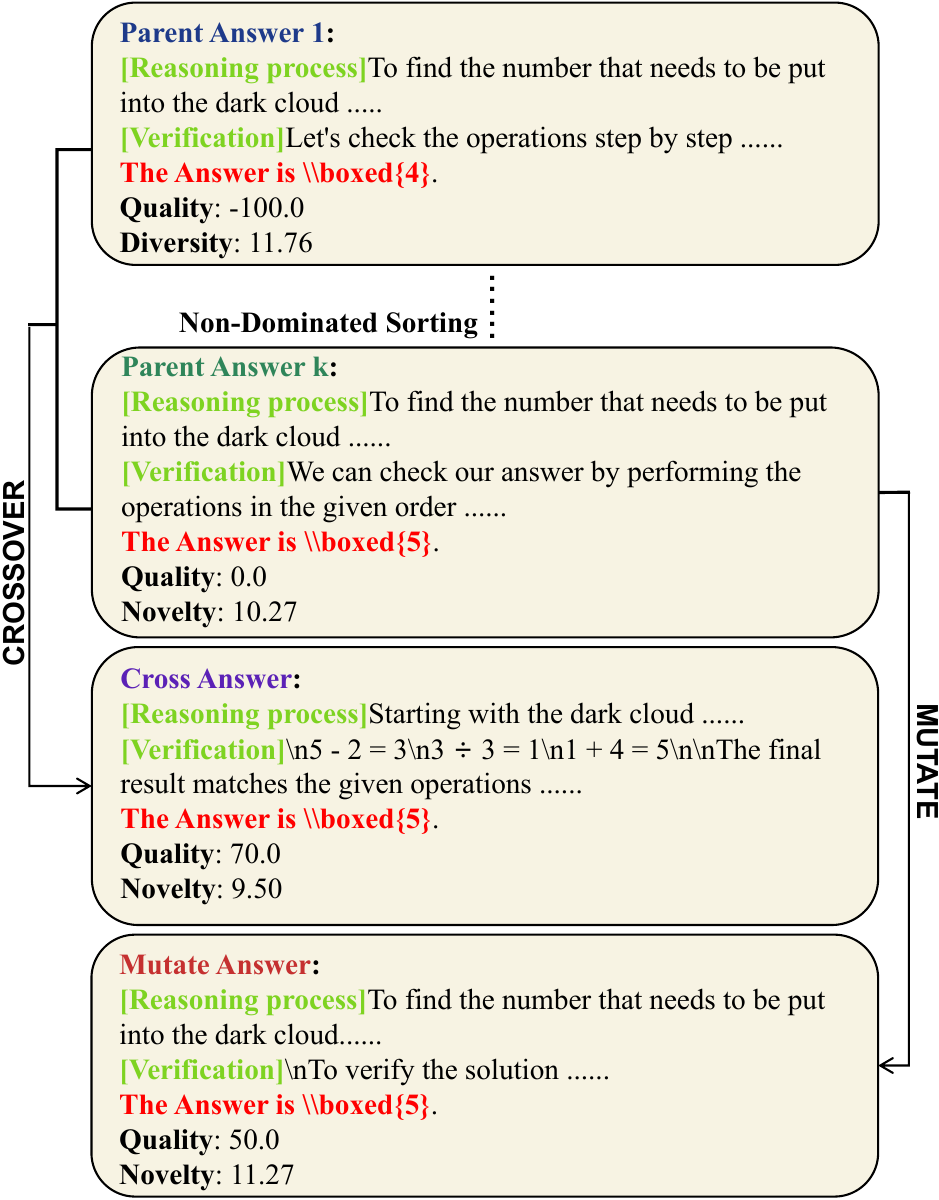}
    \label{prompts:EoT_answers}
    \vspace{-4mm}
\end{figure}

\subsection{Aggregation Answer \texorpdfstring{$\mathcal{A}^{*}$}{A*}}
\begin{figure}[H]
    \centering
    \vspace{-2mm}
    \includegraphics[width=\linewidth]{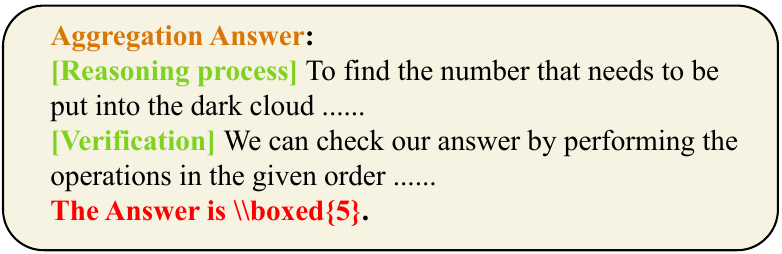}
    \label{prompts:aggregation_answer}
    \vspace{-8mm}
\end{figure}

%% file: main.bbl
\begin{thebibliography}{40}
\providecommand{\natexlab}[1]{#1}
\providecommand{\url}[1]{\texttt{#1}}
\expandafter\ifx\csname urlstyle\endcsname\relax
  \providecommand{\doi}[1]{doi: #1}\else
  \providecommand{\doi}{doi: \begingroup \urlstyle{rm}\Url}\fi

\bibitem[Ahn et~al.(2024)Ahn, Verma, Lou, Liu, Zhang, and Yin]{ahn2024large}
Janice Ahn, Rishu Verma, Renze Lou, Di Liu, Rui Zhang, and Wenpeng Yin.
\newblock Large language models for mathematical reasoning: Progresses and challenges.
\newblock \emph{arXiv preprint arXiv:2402.00157}, 2024.

\bibitem[Besta et~al.(2024)Besta, Blach, Kubicek, Gerstenberger, Podstawski, Gianinazzi, Gajda, Lehmann, Niewiadomski, Nyczyk, and Hoefler]{BestaBKGPGGLNNH24}
Maciej Besta, Nils Blach, Ales Kubicek, Robert Gerstenberger, Michal Podstawski, Lukas Gianinazzi, Joanna Gajda, Tomasz Lehmann, Hubert Niewiadomski, Piotr Nyczyk, and Torsten Hoefler.
\newblock Graph of thoughts: Solving elaborate problems with large language models.
\newblock In \emph{Conference on Artificial Intelligence}, pages 17682--17690, 2024.

\bibitem[Chen et~al.(2024)Chen, Liao, Li, and Fan]{chen2024alphamath}
Guoxin Chen, Minpeng Liao, Chengxi Li, and Kai Fan.
\newblock Alphamath almost zero: process supervision without process.
\newblock \emph{arXiv preprint arXiv:2405.03553}, 2024.

\bibitem[Cobbe et~al.(2021)Cobbe, Kosaraju, Bavarian, Chen, Jun, Kaiser, Plappert, Tworek, Hilton, Nakano, Hesse, and Schulman]{cobbe2021gsm8k}
Karl Cobbe, Vineet Kosaraju, Mohammad Bavarian, Mark Chen, Heewoo Jun, Lukasz Kaiser, Matthias Plappert, Jerry Tworek, Jacob Hilton, Reiichiro Nakano, Christopher Hesse, and John Schulman.
\newblock Training verifiers to solve math word problems.
\newblock \emph{arXiv preprint arXiv:2110.14168}, 2021.

\bibitem[Deb et~al.(2002{\natexlab{a}})Deb, Pratap, Agarwal, and Meyarivan]{996017}
K. Deb, A. Pratap, S. Agarwal, and T. Meyarivan.
\newblock A fast and elitist multiobjective genetic algorithm: Nsga-ii.
\newblock \emph{IEEE Transactions on Evolutionary Computation}, 6\penalty0 (2):\penalty0 182--197, 2002{\natexlab{a}}.

\bibitem[Deb et~al.(2002{\natexlab{b}})Deb, Pratap, Agarwal, and Meyarivan]{deb2002fast}
Kalyanmoy Deb, Amrit Pratap, Sameer Agarwal, and TAMT Meyarivan.
\newblock A fast and elitist multiobjective genetic algorithm: Nsga-ii.
\newblock \emph{IEEE transactions on evolutionary computation}, 6\penalty0 (2):\penalty0 182--197, 2002{\natexlab{b}}.

\bibitem[Du et~al.(2023)Du, Li, Torralba, Tenenbaum, and Mordatch]{du2023improving}
Yilun Du, Shuang Li, Antonio Torralba, Joshua~B Tenenbaum, and Igor Mordatch.
\newblock Improving factuality and reasoning in language models through multiagent debate.
\newblock \emph{arXiv preprint arXiv:2305.14325}, 2023.

\bibitem[Elsken et~al.(2019)Elsken, Metzen, and Hutter]{elsken2019neural}
Thomas Elsken, Jan~Hendrik Metzen, and Frank Hutter.
\newblock Neural architecture search: A survey.
\newblock \emph{Journal of Machine Learning Research}, 20\penalty0 (55):\penalty0 1--21, 2019.

\bibitem[Huang et~al.(2023)Huang, Song, Wang, Chen, and Ma]{abs-2307-10236}
Yuheng Huang, Jiayang Song, Zhijie Wang, Huaming Chen, and Lei Ma.
\newblock Look before you leap: An exploratory study of uncertainty measurement for large language models.
\newblock \emph{CoRR}, abs/2307.10236, 2023.

\bibitem[Hubert et~al.(2024)Hubert, Awa, and Zabelina]{hubert2024current}
Kent~F. Hubert, Kim~N. Awa, and Darya~L. Zabelina.
\newblock The current state of artificial intelligence generative language models is more creative than humans on divergent thinking tasks.
\newblock \emph{Scientific Reports}, 14, 2024.

\bibitem[Jones(2024)]{jones2024bigger}
Nicola Jones.
\newblock Bigger ai chatbots more inclined to spew nonsense-and people don't always realize.
\newblock \emph{Nature}, 2024.

\bibitem[Kim et~al.(2024)Kim, Kim, and Yang]{kim2024learning}
Jaehyung Kim, Dongyoung Kim, and Yiming Yang.
\newblock Learning to correct for qa reasoning with black-box llms.
\newblock \emph{arXiv preprint arXiv:2406.18695}, 2024.

\bibitem[Le~Bronnec et~al.(2024)Le~Bronnec, V{\'e}rine, Negrevergne, Chevaleyre, and Allauzen]{le2024exploring}
Florian Le~Bronnec, Alexandre V{\'e}rine, Benjamin Negrevergne, Yann Chevaleyre, and Alexandre Allauzen.
\newblock Exploring precision and recall to assess the quality and diversity of llms.
\newblock In \emph{62nd Annual Meeting of the Association for Computational Linguistics}, 2024.

\bibitem[Liang et~al.(2023)Liang, He, Jiao, Wang, Wang, Wang, Yang, Tu, and Shi]{liang2023encouraging}
Tian Liang, Zhiwei He, Wenxiang Jiao, Xing Wang, Yan Wang, Rui Wang, Yujiu Yang, Zhaopeng Tu, and Shuming Shi.
\newblock Encouraging divergent thinking in large language models through multi-agent debate.
\newblock \emph{arXiv preprint arXiv:2305.19118}, 2023.

\bibitem[Liang et~al.(2024)Liang, Song, Zheng, Wang, Yu, Li, Li, Xiong, and Li]{liang2024internal}
Xun Liang, Shichao Song, Zifan Zheng, Hanyu Wang, Qingchen Yu, Xunkai Li, Rong-Hua Li, Feiyu Xiong, and Zhiyu Li.
\newblock Internal consistency and self-feedback in large language models: A survey.
\newblock \emph{arXiv preprint arXiv:2407.14507}, 2024.

\bibitem[Liu et~al.(2024)Liu, Li, Li, Li, Zhang, Shen, and Lee]{liu2024llavanext}
Haotian Liu, Chunyuan Li, Yuheng Li, Bo Li, Yuanhan Zhang, Sheng Shen, and Yong~Jae Lee.
\newblock Llava-next: Improved reasoning, ocr, and world knowledge, 2024.

\bibitem[Lu et~al.(2024)Lu, Bansal, Xia, Liu, Li, Hajishirzi, Cheng, Chang, Galley, and Gao]{lu2024mathvista}
Pan Lu, Hritik Bansal, Tony Xia, Jiacheng Liu, Chunyuan Li, Hannaneh Hajishirzi, Hao Cheng, Kai-Wei Chang, Michel Galley, and Jianfeng Gao.
\newblock Mathvista: Evaluating mathematical reasoning of foundation models in visual contexts.
\newblock In \emph{International Conference on Learning Representations (ICLR)}, 2024.

\bibitem[Madaan et~al.(2024)Madaan, Tandon, Gupta, Hallinan, Gao, Wiegreffe, Alon, Dziri, Prabhumoye, Yang, et~al.]{Self-refine}
Aman Madaan, Niket Tandon, Prakhar Gupta, Skyler Hallinan, Luyu Gao, Sarah Wiegreffe, Uri Alon, Nouha Dziri, Shrimai Prabhumoye, Yiming Yang, et~al.
\newblock Self-refine: Iterative refinement with self-feedback.
\newblock \emph{Advances in Neural Information Processing Systems}, 36, 2024.

\bibitem[Microsoft(2024)]{microsoft2024phi35visioninstruct}
Microsoft.
\newblock Phi-3.5-vision-instruct model, 2024.

\bibitem[Newling and Fleuret(2017)]{NewlingF17}
James Newling and Fran{\c{c}}ois Fleuret.
\newblock K-medoids for k-means seeding.
\newblock In \emph{Advances in Neural Information Processing Systems}, pages 5195--5203, 2017.

\bibitem[Paul et~al.(2023)Paul, Ismayilzada, Peyrard, Borges, Bosselut, West, and Faltings]{paul2023refiner}
Debjit Paul, Mete Ismayilzada, Maxime Peyrard, Beatriz Borges, Antoine Bosselut, Robert West, and Boi Faltings.
\newblock Refiner: Reasoning feedback on intermediate representations, 2023.

\bibitem[Reimers and Gurevych(2019)]{ReimersG19}
Nils Reimers and Iryna Gurevych.
\newblock Sentence-bert: Sentence embeddings using siamese bert-networks.
\newblock In \emph{Conference on Empirical Methods in Natural Language Processing}, pages 3980--3990. Association for Computational Linguistics, 2019.

\bibitem[Schaffer(1985)]{schaffer1985multiobjective}
JD Schaffer.
\newblock Multiobjective optimization using nondominated sorting in genetic algorithms.
\newblock In \emph{Proceedings of the First International Conference on Genetic Algorithms and Their Applications}, pages 160--168. Lawrence Erlbaum Associates, 1985.

\bibitem[Shinn et~al.(2024)Shinn, Cassano, Gopinath, Narasimhan, and Yao]{shinn2024reflexion}
Noah Shinn, Federico Cassano, Ashwin Gopinath, Karthik Narasimhan, and Shunyu Yao.
\newblock Reflexion: Language agents with verbal reinforcement learning.
\newblock \emph{Advances in Neural Information Processing Systems}, 36, 2024.

\bibitem[{\'S}wiechowski et~al.(2023){\'S}wiechowski, Godlewski, Sawicki, and Ma{\'n}dziuk]{swiechowski2023monte}
Maciej {\'S}wiechowski, Konrad Godlewski, Bartosz Sawicki, and Jacek Ma{\'n}dziuk.
\newblock Monte carlo tree search: A review of recent modifications and applications.
\newblock \emph{Artificial Intelligence Review}, 56\penalty0 (3):\penalty0 2497--2562, 2023.

\bibitem[Team et~al.(2023)Team, Anil, Borgeaud, Alayrac, Yu, Soricut, Schalkwyk, Dai, Hauth, Millican, et~al.]{team2023gemini}
Gemini Team, Rohan Anil, Sebastian Borgeaud, Jean-Baptiste Alayrac, Jiahui Yu, Radu Soricut, Johan Schalkwyk, Andrew~M Dai, Anja Hauth, Katie Millican, et~al.
\newblock Gemini: a family of highly capable multimodal models.
\newblock \emph{arXiv preprint arXiv:2312.11805}, 2023.

\bibitem[Wang et~al.(2024{\natexlab{a}})Wang, Wang, Athiwaratkun, Zhang, and Zou]{wang2024mixture}
Junlin Wang, Jue Wang, Ben Athiwaratkun, Ce Zhang, and James Zou.
\newblock Mixture-of-agents enhances large language model capabilities.
\newblock \emph{arXiv preprint arXiv:2406.04692}, 2024{\natexlab{a}}.

\bibitem[Wang et~al.(2024{\natexlab{b}})Wang, Pan, Shi, Lu, Zhan, and Li]{wang2024MATHVision}
Ke Wang, Junting Pan, Weikang Shi, Zimu Lu, Mingjie Zhan, and Hongsheng Li.
\newblock Measuring multimodal mathematical reasoning with math-vision dataset, 2024{\natexlab{b}}.

\bibitem[Wang et~al.(2024{\natexlab{c}})Wang, Bai, Tan, Wang, Fan, Bai, Chen, Liu, Wang, Ge, Fan, Dang, Du, Ren, Men, Liu, Zhou, Zhou, and Lin]{Qwen2VL}
Peng Wang, Shuai Bai, Sinan Tan, Shijie Wang, Zhihao Fan, Jinze Bai, Keqin Chen, Xuejing Liu, Jialin Wang, Wenbin Ge, Yang Fan, Kai Dang, Mengfei Du, Xuancheng Ren, Rui Men, Dayiheng Liu, Chang Zhou, Jingren Zhou, and Junyang Lin.
\newblock Qwen2-vl: Enhancing vision-language model's perception of the world at any resolution.
\newblock \emph{arXiv preprint arXiv:2409.12191}, 2024{\natexlab{c}}.

\bibitem[Wang et~al.(2023)Wang, Mao, Wu, Ge, Wei, and Ji]{wang2023unleashing}
Zhenhailong Wang, Shaoguang Mao, Wenshan Wu, Tao Ge, Furu Wei, and Heng Ji.
\newblock Unleashing the emergent cognitive synergy in large language models: A task-solving agent through multi-persona self-collaboration.
\newblock \emph{arXiv preprint arXiv:2307.05300}, 2023.

\bibitem[Wei et~al.(2022)Wei, Wang, Schuurmans, Bosma, Ichter, Xia, Chi, Le, and Zhou]{Wei0SBIXCLZ22}
Jason Wei, Xuezhi Wang, Dale Schuurmans, Maarten Bosma, Brian Ichter, Fei Xia, Ed~H. Chi, Quoc~V. Le, and Denny Zhou.
\newblock Chain-of-thought prompting elicits reasoning in large language models.
\newblock In \emph{Advances in Neural Information Processing Systems}, 2022.

\bibitem[Xu(2023)]{xu2023no}
Haotian Xu.
\newblock No train still gain. unleash mathematical reasoning of large language models with monte carlo tree search guided by energy function.
\newblock \emph{arXiv preprint arXiv:2309.03224}, 2023.

\bibitem[Yang et~al.(2024)Yang, Zhang, Xu, Lu, Heng, and Lam]{yang2024unveiling}
Haoran Yang, Yumeng Zhang, Jiaqi Xu, Hongyuan Lu, Pheng~Ann Heng, and Wai Lam.
\newblock Unveiling the generalization power of fine-tuned large language models.
\newblock \emph{arXiv preprint arXiv:2403.09162}, 2024.

\bibitem[Yang et~al.(2023)Yang, Li, Lin, Wang, Lin, Liu, and Wang]{yang2023dawn}
Zhengyuan Yang, Linjie Li, Kevin Lin, Jianfeng Wang, Chung-Ching Lin, Zicheng Liu, and Lijuan Wang.
\newblock The dawn of lmms: Preliminary explorations with gpt-4v (ision).
\newblock \emph{arXiv preprint arXiv:2309.17421}, 9\penalty0 (1):\penalty0 1, 2023.

\bibitem[Yao et~al.(2022)Yao, Zhao, Yu, Du, Shafran, Narasimhan, and Cao]{yao2022react}
Shunyu Yao, Jeffrey Zhao, Dian Yu, Nan Du, Izhak Shafran, Karthik Narasimhan, and Yuan Cao.
\newblock React: Synergizing reasoning and acting in language models.
\newblock \emph{arXiv preprint arXiv:2210.03629}, 2022.

\bibitem[Yao et~al.(2023)Yao, Yu, Zhao, Shafran, Griffiths, Cao, and Narasimhan]{YaoYZS00N23}
Shunyu Yao, Dian Yu, Jeffrey Zhao, Izhak Shafran, Tom Griffiths, Yuan Cao, and Karthik Narasimhan.
\newblock Tree of thoughts: Deliberate problem solving with large language models.
\newblock In \emph{Advances in Neural Information Processing Systems}, 2023.

\bibitem[Ye et~al.(2024)Ye, Yang, Pang, Wang, Wong, Yilmaz, Shi, and Tu]{abs-2401-12794}
Fanghua Ye, Mingming Yang, Jianhui Pang, Longyue Wang, Derek~F. Wong, Emine Yilmaz, Shuming Shi, and Zhaopeng Tu.
\newblock Benchmarking llms via uncertainty quantification.
\newblock \emph{CoRR}, abs/2401.12794, 2024.

\bibitem[Zhang et~al.(2024)Zhang, Huang, Zhou, Li, and Ouyang]{abs-2406-07394}
Di Zhang, Xiaoshui Huang, Dongzhan Zhou, Yuqiang Li, and Wanli Ouyang.
\newblock Accessing {GPT-4} level mathematical olympiad solutions via monte carlo tree self-refine with llama-3 8b.
\newblock \emph{CoRR}, abs/2406.07394, 2024.

\bibitem[Zhao et~al.(2024)Zhao, Zhang, Yu, Fei, Huang, and Yang]{zhao2024many}
Jiale Zhao, Huijie Zhang, Huanhuan Yu, Hansheng Fei, Xiangdang Huang, and Qiuling Yang.
\newblock A many-objective evolutionary algorithm based on three states for solving many-objective optimization problem.
\newblock \emph{Scientific Reports}, 14\penalty0 (1):\penalty0 19140, 2024.

\bibitem[Zhou et~al.(2024)Zhou, Liu, Zhao, and Gong]{Zhou0Z024}
Zhipeng Zhou, Liu Liu, Peilin Zhao, and Wei Gong.
\newblock Pareto deep long-tailed recognition: {A} conflict-averse solution.
\newblock In \emph{International Conference on Learning Representations}, 2024.

\end{thebibliography}
